\documentclass[conference]{ieeeconf}
\usepackage{xparse}
\usepackage{ifthen}
\newboolean{anonymous}
\setboolean{anonymous}{true}  
\usepackage{graphicx}
\usepackage{amsfonts}
\usepackage{amsmath}
\usepackage{subcaption}

\NewDocumentCommand{\acite}{m}{%
    \ifthenelse{\boolean{anonymous}}
    {[Anonymous]}  
    {\cite{#1}}    
}

\NewDocumentCommand{\aauthor}{m}{%
    \ifthenelse{\boolean{anonymous}}
    {Anonymous}  
    {#1}        
}

\newcommand{\corr}[1]{{\color{black}#1}}

\usepackage{booktabs}
\usepackage{multirow}
\usepackage{bold-extra}

\usepackage{times}

\usepackage{multicol}
\usepackage[bookmarks=true]{hyperref}
\usepackage{color}
\usepackage[ruled,vlined]{algorithm2e}
\usepackage{caption}
\usepackage{cuted}
\usepackage{cite}
\usepackage{hyperref}

\IEEEoverridecommandlockouts                              
\begin{document}

\title{PDDL-ART: Autonomous Symbolic Abstraction From Demonstration For Long-Horizon Robotic Manipulation Using Vision-Language Models}

\author{Disha Kamale and Dmitry Berenson
\thanks{Disha Kamale and Dmitry Berenson are with the Robotics Department,
UNiversity of Michigan, Ann Arbor, MI, 48109
{\tt\small \{dkamale, dmitryb\}@umich.edu}}%
}



%

\maketitle

\IEEEpeerreviewmaketitle


\begin{abstract}
 Symbolic planning with PDDL offers a principled framework for long-horizon robot manipulation, but constructing accurate PDDL domain and problem descriptions remains a significant bottleneck, typically requiring substantial domain expertise. We present a Vision-Language Model (VLM)-based approach called \textsc{PDDL-ART}, a framework that autonomously generates task-specific PDDL domain and problem descriptions from a single expert demonstration, a natural language task description, and a library of available high-level action names. PDDL-ART does not require any domain templates, action signatures, or fine-tuning. To ensure the generated descriptions are not only syntactically valid but semantically aligned with the demonstrated task, \textsc{PDDL-ART} introduces a multi-stage correction pipeline operating at syntactic, semantic, and execution levels. A key component of execution-guided correction is symbolic predicate grounding. Instead of relying solely on visual observations, PDDL-ART leverages the tool-use capabilities of modern VLMs to incorporate geometric and temporal reasoning for evaluating relational predicates that are not directly discernible from images alone. Critically, the model autonomously determines when to invoke these tools and how to interpret their outputs. We evaluate PDDL-ART on challenging manipulation tasks in engine maintenance and household domains, including tasks that require memory, abstract predicate inference, and goal states that are visually indistinguishable from the initial state. \textsc{PDDL-ART} achieves an average success rate of 93.3\%, compared to 78.3\% for a VLM-based planner. 
 \end{abstract}

\section{Introduction}



Long-horizon robot manipulation tasks require reasoning over actions, objects, and the causal dependencies between them. Consider the task of autonomously changing engine oil: the robot must first check the oil level using a dipstick, drain the old oil, and replace the oil filter before refilling. Each step depends on the successful completion of the previous one. Such dependencies make planning challenging since the correctness of any action may depend on intermediate states produced by actions executed earlier in the sequence.

Symbolic planning methods provide a structured way to represent and solve such problems. Classical planners can synthesize long-horizon action sequences with guarantees such as soundness and completeness with respect to the provided model. However, these methods require an accurate symbolic abstraction of the problem. Constructing these abstractions manually requires domain expertise and can be time consuming. In addition, the hand-crafted abstractions need to be modified when the task, objects, environment or robot capabilities change.

To mitigate these limitations, prior work has explored methods for learning relational symbolic abstractions from robot data. These methods infer predicates, symbolic states, and action models from demonstrations, kinematic trajectories, interaction data, or geometric relations \cite{LAMP, migimatsu2022grounding}. These approaches reduce the manual specification required. However, learned abstractions can be difficult to inspect and revise. For instance, when planning fails, it may not be clear whether the error arises from an incorrect predicate, grounding, or action sequencing. This issue becomes more pronounced in long-horizon tasks, where such errors in logical states can lead to invalid plans.

 Recently, there has been a growing interest in studying the use of Large Language Models (LLMs) and Vision-Language Models (VLMs) for generating planning descriptions from language and visual inputs~\cite{liu2023llm+, adamik2025planowl, ye2025pretraining, yan2025using, yang2024guiding}. These models can be used to propose predicates, action descriptions, and formalisms without task-specific fine-tuning. In this work, we consider the problem of generating symbolic abstractions from a single expert demonstration and natural language task description. We focus on abstractions represented as PDDL - a standard formalism for encoding deterministic planning problems that involve long-horizon reasoning. However, generating PDDL descriptions without interacting with the environment and checking downstream feasibility remains error-prone. The output descriptions must not only be syntactically correct, but must also define a feasible planning problem, infer initial and goal states that are aligned with the demonstration, and produce plans that are executable on the robot. If any one of these components is incorrect, the generated PDDL description may be invalid, infeasible, or may encode a problem different from the intended task.

\begin{figure}
    \centering
    \includegraphics[width=\linewidth]{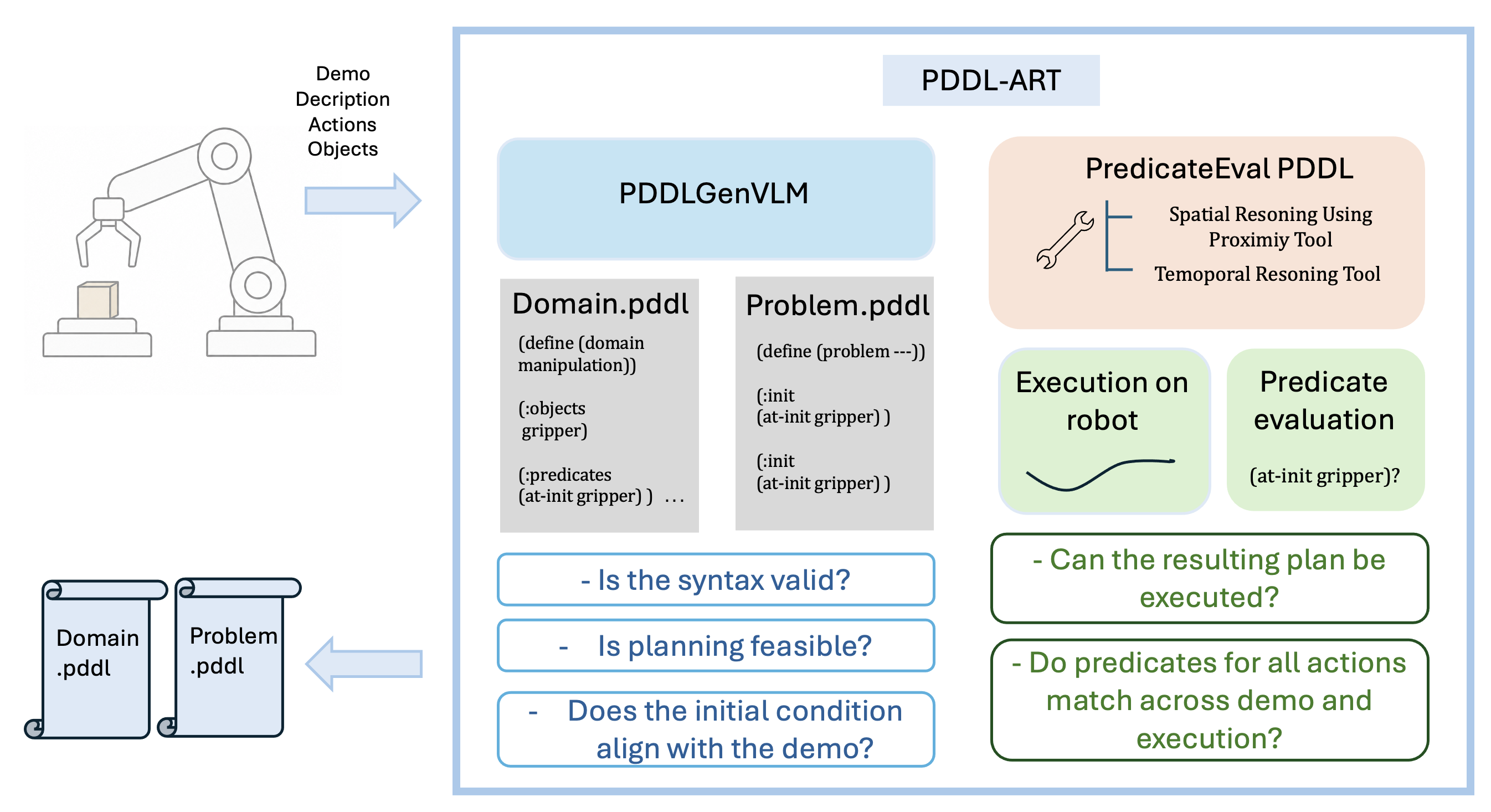}
    \caption{\small The proposed PDDL-ART framework. Given a single expert demonstration, task description, library of available action names and objects, PDDL-ART produces PDDL domain and problem files that are validates at syntactic, semantic, execution, and predicate evaluation level. }
    \label{fig:placeholder}
\end{figure}

We propose PDDL-ART (PDDL generation with Automated Reasoning and Tool use), a framework for generating task-specific PDDL domain and problem files using VLMs. Our method takes an input a single expert demonstration, natural language minimal task description, and a library of available high-level actions. and generates the necessary predicates, selects actions and defines action schemas, parameter types, preconditions, and effects, as well as initial and goal conditions required for planning. Note that, we do not assume a manually provided template, ground-truth action signatures, or candidate predicates. PDDL-ART instead constructs the PDDL domain and problem files directly from the input information. 

In order to ensure that the generated PDDL descriptions are not only syntactically valid but also aligned with the task demonstration, PDDL-ART includes a multi-stage validation and correction pipeline. The generated PDDL is checked for syntactic validity, semantic consistency, and execution-level correctness. When failure is detected, the VLM is re-queried with the generated PDDL descriptions and the error message for failure to guide correction. \corr{To achieve this, PDDL-ART consists of two components: 1) PDDLGenVLM - generates PDDL domain and problem files and systematically validates their syntax and semantics, 2) PredicateEvalVLM - performs execution-guided correction to ensure that the generated plan is executable on a robot, and the generated lan aligns with the task given in demonstration. }

An important part of execution-guided correction is symbolic predicate grounding. PDDL-ART grounds predicate evaluation using visual observations together with geometric and temporal reasoning. To evaluate the logical states for predicates for which visual data alone may be insufficient, PredicateEvalVLM uses the tool-use capabilities of modern VLMs to aid in spatial reasoning as well as keeping track of past logical states. It is important to note that the model autonomously determines when to invoke these tools, and how to interpret their outputs. The resulting output of our method is thus \corr{systematically-validated PDDL problem and domain descriptions that represent the demonstrated task}. 

Our key contributions are: 
1) We introduce a multi-stage correction framework that systematically validates and rectifies the generated PDDL descriptions at syntactic, semantic as well as execution levels. 
2) We introduce a novel execution-guided cross-validation approach for correcting the generated PDDL descriptions. The proposed predicate evaluation method grounds symbolic predicate evaluation in geometric and temporal reasoning. This allows the framework to evaluate relational predicates that are not immediately discernible from visual observations. 
3) We demonstrate the performance of PDDL-ART on tasks that require memory, inventing abstract predicates, as well as problems where the goal states are visually indistinguishable from the initial states, in engine maintenance and household domains in simulation.  PDDL-ART achieves an average demo-aligned success rate of \corr{ 93.3\%, whereas the best baseline achieves a success rate of 78.3\%. }

The codebase and prompts for PDDL-ART will be made available open source upon publication.

\section{Related Work}

\begin{table}[]
\fontsize{7}{8.5}\selectfont
\begin{minipage}{\columnwidth}   
\begin{tabular}{l|l|l|l|l}
 Methods & Domain & Problem & No Expert & No Fine-tuning \\ \hline
VLMFP \cite{hao2025simulation} & \checkmark  & \checkmark &  \checkmark & $\times$  \\
LLMs world-models \cite{guan2023leveraging} & \checkmark & \checkmark    & $\times$  & \checkmark \\
VLLM formalizer \cite{he2025vision} & $\times$ & \checkmark  & \checkmark  & \checkmark \\
PDDL-ART (Ours)              & \checkmark & \checkmark    & \checkmark & \checkmark \\
\end{tabular}
\end{minipage}
\caption{Comparison of closely related works.}
\label{tab:related}
\end{table}

\subsubsection{LLMs and VLMs as planners} 
LLMs and VLMs have been widely applied to planning tasks, leveraging their commonsense reasoning capabilities for generating subgoals~\cite{yang2024guiding}, guiding task and motion planning~\cite{zhang2025llm}, and incorporating logical constraints~\cite{pan2023logic, yan2025using, balaji2025language}. Recent work has also explored using LLMs to translate between formal representations and natural language as an intermediary for planning~\cite{verma2025teaching}. However, while LLMs and VLMs perform well on text-based planning benchmarks, their limited spatial understanding remains a significant challenge for physical, real-world planning tasks~\cite{wu2022autoformalization}.

\subsubsection{Foundation Models as Formalizers}

This research direction leverages the reasoning capabilities of foundation models to produce structured formal representations instead of directly generating plans. Classical planners can operate on such abstractions to synthesize plans~\cite{wu2022autoformalization, hao2025simulation, silver2023predicate, tantakoun2025llms, athalye2026pixels}. For instance,~\cite{liu2024lang2ltl} grounds navigation commands to temporal logic representations using VLMs. Among these, PDDL has emerged as a widely adopted target representation given its ability to precisely encode long-horizon symbolic planning tasks. LLM + P~\cite{liu2023llm+} generates a PDDL problem file from natural language, solves it with a classical planner, and translates the plan back to natural language. NL2PLAN~\cite{mahdavi2024leveraging} iteratively constructs both domain and problem files, though it requires ground truth action signatures as input.~\cite{he2025vision} and~\cite{aregbede2025here} assume the domain file is provided and focus solely on problem file generation.~\cite{athalye2026pixels} and~\cite{silver2023predicate} explore predicate invention in the context of bilevel planning. UniDomain~\cite{ye2025pretraining} pretrains a generalised PDDL domain for robotic tasks by composing actions, preconditions, and effects from a dataset of demonstrations. Table~\ref{tab:related} shows a comparison of our method with closely related works. 

The most closely related work is~\cite{hao2025simulation}, which similarly uses execution-guided correction to autonomously generate PDDL domain and problem files. However, their framework requires fine-tuning a VLM on domain-specific data to embed spatial reasoning. In contrast, our framework embeds geometric and temporal reasoning directly through tool use, avoiding domain-specific fine-tuning while remaining applicable to unseen domains.

\section{Preliminaries}

An automated planning problem is a goal-directed deterministic planning problem: given a description of the world and a desired outcome, the task is to synthesize a sequence of high-level actions that transitions the world from an initial state to one satisfying the goal, while respecting the constraints of the environment.

\subsection{PDDL: Planning Domain Definition Language}

Planning Domain Definition Language (PDDL)~\cite{aeronautiques1998pddl} is the standard formalism for encoding such planning tasks.
A PDDL task consists of two components: 1) a \textit{domain} that describes the \corr{properties of the world}, what actions are available to the robot, etc., and 2) a \textit{problem instance} that describes the specifics of a given scenario.  
Formally, a PDDL task is a pair $\langle D, P \rangle$ comprising a domain $D$ and a problem instance $P$.

\subsubsection{Domain}
The domain $D = \langle \mathcal{T}, \mathcal{P}, \mathcal{A} \rangle$ encodes the general rules of the environment, where $\mathcal{T}$ is a type hierarchy, $\mathcal{P}$ a set of predicates, and $\mathcal{A}$ a set of action schemas.
The type hierarchy $\mathcal{T}$ defines the categories of objects that can exist in the world and their inheritance relationships.
Predicates $\mathcal{P}$ form the vocabulary for describing world properties: each predicate $p \in \mathcal{P}$ is declared over a tuple of typed arguments, with each argument assigned a type $t \in \mathcal{T}$.

In their declared form, predicates and actions are \emph{lifted} i.e., they contain uninstantiated variables rather than concrete objects.
We denote lifted predicates and actions as $\bar{p}$ and $\bar{a}$, respectively.
Substituting all variables with concrete objects $o \in \mathcal{O}$ of compatible types instantiates them into their \emph{grounded} form, denoted $p$ and $a$.
Grounding predicate $\bar{p}$ yields a Boolean proposition $p(o_1, \ldots, o_m)$ that denotes a specific property of the world that can be evaluated as $\mathrm{true}$ or $\mathrm{false}$.

Each action schema $\bar{a} \in \mathcal{A}$ is a tuple, $    \bar{a} = \langle \mathit{name}(\bar{a}),\ \mathit{pre}(\bar{a}),\ \mathit{eff}(\bar{a}) \rangle$
where $\mathit{name}(\bar{a})$ is its identifier, $\mathit{pre}(\bar{a})$ is its \emph{precondition} - a first-order logical formula over $\mathcal{P}$ specifying which properties must hold for the action to be executable and $\mathit{eff}(\bar{a})$ is its \emph{effect}, defining how the state of the world changes upon execution.

\subsubsection{Problem}
The problem instance $P = \langle \mathcal{O}, I, G \rangle$ defines the specifics of the task at hand.
The object set $\mathcal{O}$ contains the concrete entities that populate the scenario, where each object $o_k = \langle n, t \rangle$ has a name $n$ and a type $t \in \mathcal{T}$.
The initial state $I$ assigns a Boolean truth value to every grounded proposition, fully characterising the state of the world at the outset.
The goal $G$ is a logical formula over $\mathcal{P}$ and $\mathcal{O}$ characterising the set of desired world states.

\subsubsection{Plans}
A plan $\pi = \langle a_1, \ldots, a_m \rangle$ is a sequence of $m$ grounded actions such that each $a_i$ is executable in the state produced by executing $a_1, \ldots, a_{i-1}$ from $I$, and the resulting final state satisfies $G$.
Synthesizing such plans is the task of a \emph{symbolic planner}, which takes the domain and problem instance as input and returns a valid plan $\pi$, if it exists,  or planning failure, which we denote using $\bot$.

In this work, we consider a subset of typed PDDL with negative preconditions, sufficient for describing the engine domain and household tasks considered for evaluation.

\noindent \textbf{Notation.}
The pose of an object $A$ with respect to $B$ is denoted as $\mathbf{T}_A^B \in SE(3)$.  For a trajectory $\mathbf{x} = \{x_t\}_{t=0}^{T}$, $\mathbf{x}_{[a:b]}$ denotes the subsequence from $a$ to $b$.

\begin{figure*}
\centering
  \includegraphics[width=.7\linewidth]{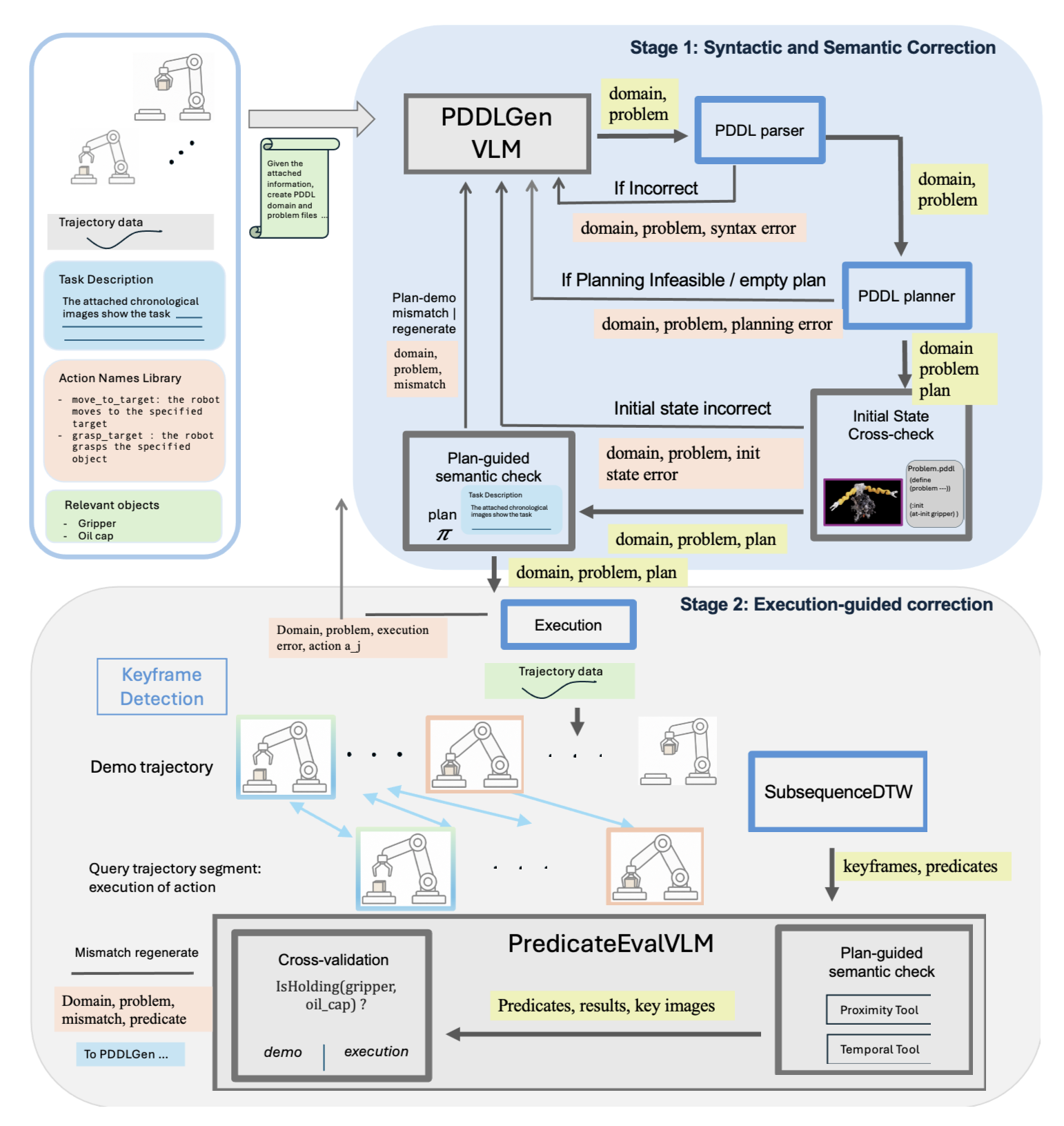}
\caption{\footnotesize PDDL-ART Overview: The framework takes as input a single expert demonstration, a natural language task description, a high-level action library, and a list of relevant objects. Stage 1: PDDLGen outputs PDDL domain and problem files that are syntactically and semantically validated. If a valid plan is found, Stage 2 performs execution-guided corrections, cross-validating the generated PDDL against demo. The framework thus outputs empirically validated PDDL domain and problem files. All rectangles with bold grey outlines indicate a VLM component. }
\label{fig:pddlart_stage_1}
\vspace{-4pt}
\end{figure*}

\section{Problem Setting}
The problem of demo-guided PDDL generation has four main input components described below. 

\subsection{Expert Demonstration} 
The primary input to our framework is a task demonstration $\mathcal{D}$ 
collected by executing an expert plan on the task. These can be plans generated using classical planners, roll-outs of a learned expert policy, or the recorded data from a human expert performing the task via teleoperation. A demonstration consists of a state trajectory $\zeta$ and a set of RGB images $\mathcal{V}$. Formally, the state trajectory of $T$ timesteps,
\begin{equation}
    \zeta = \{s_t\}_{t=1}^{T},
\end{equation}
where each state $s_t$ comprises of the robot end-effector pose 
$\mathbf{e}_t \in SE(3)$, the gripper state $g_t \in \{0,1\}$, 
and the poses of all task-relevant objects $\mathcal{O}_{task} = 
\{o_1, \ldots, o_K\}$. 

The demonstration is accompanied by a sequence of images $\mathcal{V} = \{v_t\}_{t=1}^{T}$ corresponding to each timestep of task.

\subsection{Natural Language Task Description}
The natural language task description $d$ provides a structured summary of the task to be performed. Concretely, $d$ encodes: 1) a brief description of the overall task and a brief summary of how the task is performed; 2) the description of the robot initial state with respect to the 
    task-relevant objects (for instance, whether the gripper is holding a mug at the beginning of the task); 3) the set of task-relevant objects $\mathcal{O}_{task}$; and, 4) the verbal description of the success criteria.

\subsection{High-level Action Library}
The action name library $\mathbb{A} = \{a^1, a^2, \ldots, a^M\}$ defines a vocabulary of $M$ high-level action names available to the robot. Unlike fully-specified action schemas, the library provides only the action identifiers. Our method is responsible for inferring all formal properties of each action, including its parameter types $t \in \mathcal{T}$, arity, the necessary predicates set $\mathcal{P}$, and the preconditions $\mathit{pre}(\cdot)$ and effects $\mathit{eff}(\cdot)$ required to produce a valid 
PDDL action schema.

\subsection{Problem Definition}
\label{subsec:problem}
Given an input tuple $(\mathcal{D}, d, \mathbb{A}, \mathcal{O}_{task})$, our aim is to generate a PDDL task description $\langle D, P \rangle$ such that a downstream classical planner finds a valid plan $\pi$ that is semantically consistent with the demonstrated task.


\section{Method}

Our framework addresses the problem of automated PDDL generation from expert demonstration through a multi-stage correction and systematic validation pipeline. The proposed framework consists of two complementary VLM components: PDDLGen, which generates and refines the PDDL domain and problem files, and PredicateEvalVLM, which validates the correctness of the generated action schemas against the demonstration. The PDDL-ART framework is outlined in Figure~\ref{fig:pddlart_stage_1}\footnote{Example prompt structures at various stages of PDDL-ART are available at https://anonymous.4open.science/w/prompts-4E82/}. 

The pipeline proceeds in two stages. Stage 1 focuses entirely on correcting the generated PDDL files prior to any robot execution, verifying their syntactic correctness, planning feasibility, and semantic consistency with the task. Stage 2 executes the resulting plan on the robot, using visual and geometric verification of each action's preconditions and effects against the demonstration to catch any remaining errors, with execution failures fed back as correction signals.

Note that, throughout both stages, the {corrections are only queried minimally. PDDLGen receives specific error feedback and is prompted to create minimal modifications sufficient to resolve the identified issue.} This aims to preserves the valid structure accumulated across correction rounds and prevents the model from discarding correct components. We now present the detailed description of each component of the proposed framework.

\subsection{Stage-1: Syntactic and Semantic Correction} In this stage, {PDDLGen} verifies whether the generated PDDL domain and problem files are syntactically and semantically valid before proceeding to plan execution. Correction is applied iteratively: if a check fails, the corresponding error or mismatch message is appended to the correction prompt and PDDLGen is queried to regenerate the files.  {We set a maximum of one correction round per check, however, multiple iterations could be used if desired}. 

Formally, this stage takes the input $(\mathcal{D}, d, \mathbb{A}, \mathcal{O}_{task})$. For long-horizon tasks, passing the full frame 
sequence $\mathcal{V}$ to the VLM is impractical as redundant frames inflate context length and may affect the model's inference accuracy. Therefore, we extract a subset $\mathcal{I} \subset \mathcal{V}$ of keyframes via a simple proximity-based heuristic, where each $v_k \in \mathcal{I}$ corresponds to a timestep at which an important state transition occurs in the trajectory $\zeta$.

\subsubsection{Syntactic Check} The first check verifies syntactic correctness using an off-the-shelf PDDL parser (VAL~\cite{val}). A PDDL file is syntactically valid if it conforms to the PDDL grammar - for instance, all predicates referenced in action schemas must be declared in the \texttt{(:predicates)} section, and all types assigned to objects must appear in the \texttt{(:types)} hierarchy. If parsing fails, the parser error is fed back to {PDDLGen} for correction. Only files that pass the syntactic check proceed to the semantic checks below. 

\subsubsection{Planning Feasibility Check} A syntactically valid PDDL description may still be semantically ill-formed if no valid plan exists. We submit the generated domain and problem files to the Fast Downward planner~\cite{helmert2006fast} and identify two failure modes: (a) planning failure ($\bot$), where the planner cannot find any plan, indicating an error in the domain, problem, or both; and (b) empty plans, where the goal $G$ is trivially satisfied in the initial state $I$, indicating that the initial and/or goal conditions in the problem file are incorrectly specified. In either case, the planner output is returned to {PDDLGen} for correction. \subsubsection{Initial State Cross-Check} Since the initial state $I$ is inferred by the model rather than provided as input, we perform a visual cross-validation to verify its correctness. Concretely, we prompt the VLM with the initial frame $v_0 \in \mathcal{V}$ of the demonstration alongside the generated problem file, and ask it to identify any mismatches between the observed scene and the declared initial state. Any detected inconsistency is returned to {PDDLGen} as a structured mismatch message for correction. 

\subsubsection{Plan-Guided Semantic Check} The last step in stage-1 verifies that the generated plan $\pi_k$ produced from the most recent PDDL files $D^k$ and $P^k$, where $k$ is the number of correction iterations the files have gone through up until this stage, is semantically consistent with the intended task. We prompt the VLM with the plan $\pi_k$ and the task description $d$, and ask it to identify any misalignments between the two, returning the mismatches to PDDLGen for correction.

The final check verifies that the generated plan $\pi_k$, produced from the corrected PDDL files $\langle D^k, P^k \rangle$ after $k$ correction iterations, is semantically consistent with the intended task. We prompt the VLM with $\pi_k$ and the task description $d$, and ask it to identify any misalignments between the two. Any identified misalignments are returned to {PDDLGen} for a final round of correction before proceeding to the next stage of execution-guided corrections.

Note that, in the absence of an expert in the loop and given the minimal task description provided as input, formally verifying whether the generated PDDL and the demonstration encode the same underlying task is {a challenging problem}. This check therefore serves as a proxy for catching semantic drift between the generated PDDL and the intended task.

\subsection{Stage-2: Execution-guided correction} 

The generated plan $\pi^k = \langle a_1, \ldots, a_m \rangle$ is executed on the robot via a refinement function $\Lambda : \mathcal{A} \rightarrow \mathcal{Q}$, which maps each grounded high-level action $a_i$ to the sequence of robot configurations required for its physical execution, where $\mathcal{Q}$ denotes the robot configuration space. If execution fails at action $a_j$, for instance due to a physically infeasible action such as \texttt{pick\_up(apple)} when the gripper is already holding an object, we use the failure as feedback to the VLM to correct the PDDL files. Specfically, this stage passes the failed action $a_j$, the successfully executed plan prefix $\langle a_1, \ldots, a_{j-1} \rangle$, the current PDDL files $\langle D^k, P^k \rangle$, and the full plan $\pi^k$ to the VLM to identify the cause of failure and apply a minimal correction to the PDDL files.

If no execution failure occurs, we proceed to the final validation stage, where PredicateEvalVLM cross-validates the preconditions $\mathit{pre}(a_j)$ and effects $\mathit{eff}(a_j)$ of each action $a_j \in \pi^k$ against the demonstration $\mathcal{D}$. Since we do not assume privileged access to the underlying high-level action segmentation of the expert demonstration, we resort to subsequence matching to localize the demonstration segment that corresponds to each action execution.

\subsubsection{Keyframe Detection via Relational Trajectory Matching} For each grounded action $a_j \in \pi^k$ with relevant object $o_j$, we compute the end-effector trajectory expressed relative to the initial pose of $o_j$. Formally, let $\mathbf{e}^{o_{j,0}}_{[0:T]}$ denote the full demonstration end-effector trajectory in the reference frame of $o_j$ at its initial pose, and let $\hat{\mathbf{e}}^{o_{j,0}}_{[a':b']}$ denote the corresponding relative trajectory recorded during the execution of $a_j$ on the robot. We identify the demonstration subsequence most similar to the executed action trajectory via Dynamic Time Warping (DTW) Subsequence Matching~\cite{memmel2025strap}, 

{
\begin{equation} 
\label{eq:dtw}
\mathbf{e}^{o_{j,0}}_{[a:b]} = \mathrm{SubsequenceDTW}\!\left( \mathbf{e}^{o_{j,0}}_{[0:T]},\, \hat{\mathbf{e}}^{o_{j,0}}_{[a':b']} \right), 
\end{equation} 
}
\noindent where $[a, b] \subseteq [0, T]$ is the identified matching interval in the demonstration. The RGB frames $v_a$ and $v_b$ at the boundaries of this interval serve as keyframes capturing the world state immediately before and after the execution of $a_j$, respectively.

\corr{ Matching is performed one action at a time. For each executed action $a_j$, subsequence matching identifies the segment of the demonstration trajectory most similar to the execution trajectory of $a_j$. If a match is found, it implies that the demonstration and execution are performing similar actions and thus, the predicates corresponding to the preconditions and effects of $a_j$ can be cross-validated across both demo and exeution instances. If predicate cross-validation results match across all execution actions, it suggests that the generated PDDL encodes the same task as the demonstration.  }

\subsubsection{Predicate evaluation-guided correction}
Given the identified demo and available execution keyframes, PredicateEvalVLM independently evaluates each predicate from $\mathit{pre}(a_j)$ and $\mathit{eff}(a_j)$ on both demo and execution keyframes. Note that, pure visual inspection may often be insufficient for complex manipulation problems, where the predicates may often involve relational and geometric information, as well as abstract symbolic states. To address this, we leverage the tool-use capabilities of modern VLMs where the model has access to the following tools: 

1) Proximity Inspection Tool: evaluates geometric 
and relational properties of the scene, such as \texttt{IsHolding (gripper, nut)}. The model determines which predicates require the geometric evaluation,   ii  autonomously infers the input arguments and appropriate thresholds to use for proximity-based checks. \corr{This is achieved by prompting the model to reason about what the proximity threshold would be appropriate for a given grounded predicate instance. }

2) Temporal Tool: retrieves the most recent truth value of a logical state as execution evolves. This tool becomes especially important for longer horizon tasks, where the truth value of a predicate at the current timestep may be determined by some action executed much earlier in the plan. 

Note that the model autonomously decides whether to invoke a tool, which tool may be approriate for a given predicate and what parameters to use when using a tool. 

The result of predicate evaluations is passed to the next layer of the validator which cross-validates the results for consistency and provides a verdict. If there is a mismatch in predicate evaluation results for demo and execution instances, the PredicateEvalVLM also provides a brief reason for the mismatch. This output, combined with the most recent domain, problem files as well as the action $a_j$ and the set of predicates for which mismatch was observed, are then passed back to PDDLGen for final correction. The resulting PDDL files are considered the final output of our framework. If no mismatch is found, this stage outputs the most recent PDDL files verbatim.

\begin{algorithm}
\caption{\textit{PDDL-ART}() }
\label{alg:pddlart}
\footnotesize
\KwIn{Demonstration $\mathcal{D}$, task description $d$, 
      action library $\mathbb{A}$, objects $\mathcal{O}$}
\KwOut{Domain $D$, Problem $P$}
\DontPrintSemicolon
\BlankLine

\tcp{\footnotesize Initial Generation}
$\langle D, P \rangle \leftarrow \text{PDDLGen}(\mathcal{D}, d, \mathbb{A}, \mathcal{O})$ \;
\BlankLine

\tcp{\footnotesize Stage 1: Syntactic and Semantic Correction}
$\langle D, P \rangle \leftarrow \text{SyntaxCorrect}(D, P)$ \;
$\pi, \langle D, P \rangle \leftarrow \text{FeasibilityCorrect}(D, P)$ \;
$\langle D, P \rangle \leftarrow \text{InitStateCorrect}(D, P, v_0)$ \;
$\pi, \langle D, P \rangle \leftarrow \text{PlanSemanticCorrect}(\pi, D, P, d)$ \;
\BlankLine

\tcp{\footnotesize Stage 2: Execution-Guided Correction}
$j \leftarrow \text{Execute}(\pi, \Lambda)$ \;
\If{$j == \bot$}{
    $\langle D, P \rangle \leftarrow \text{PDDLGen}(\langle a_1, \ldots, a_{j-1} \rangle, a_j, D, P)$ \;
    $\pi \leftarrow \text{Planner}(D, P)$ \;
    $\text{Execute}(\pi, \Lambda)$ \;
}
    \For{$a_j \in \pi$}{
        $[a, b] \leftarrow \mathrm{SubsequenceDTW}\!\left( \mathbf{e}^{o_{j,0}}_{[0:T]},\, \hat{\mathbf{e}}^{o_{j,0}}_{[a':b']} \right), $ \hfill \small eq.(\ref{eq:dtw}) \;
        $\langle D, P \rangle \leftarrow \text{PredicateEvalVLM}(v_a, v_b, a_j, D, P)$ \;
    
    \If{\texttt{mismatch}}{
        $\langle D, P \rangle \leftarrow \text{PDDLGen}(\langle a_1, \ldots, a_{j-1} \rangle, a_j, pre(a_j), eff(a_j), D, P)$ \;
        break \; 
    } }
\BlankLine
\Return $\langle D, P \rangle, \pi$ \;

\end{algorithm}

\section{Experiments}

\begin{figure*}[t]
  \centering


\includegraphics[width=\linewidth, ]{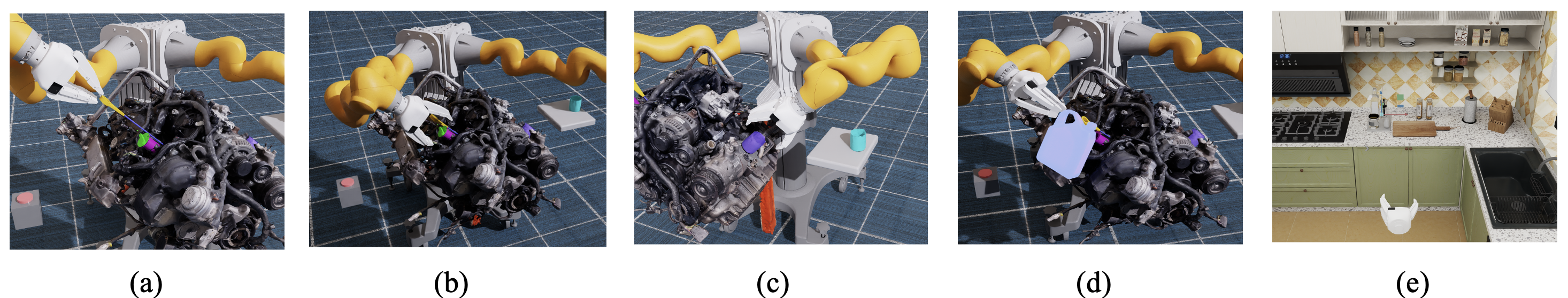}

  \caption{ \small
Case study scenarios. (a) Task 1: Check oil levels, (b) Task 2: Oil cap manipulation, 
 (c) Task 3: Change oil filter, 
 (d) Task 4: Change oil,
 (e) Task 5 and 6: Cooking Prep I, II.
}
  \label{fig:experiments}
\end{figure*}

We test the proposed PDDL-ART framework on six task scenarios, two are household domain tasks inspired by standard task planning benchmarks, and four are scenarios related to engine maintenance. These task scenarios range from 2 to 12 objects and 5 to 37 actions. Note that, we have intentionally chosen tasks that require memory and may require abstract predicates since the start and goal state for the object looks the same. For instance, at the beginning of oil change task, the cap is fastened on the engine, implying that the initial condition should have a predicate \texttt{IsFastened cap}. At the end of the task, the robot is expected to screw the cap back on and the goal thus involve the same predicate. To successfully generate PDDL problem, PDDLGenVLM needs to infer that an abstract predicate \texttt{IsOpenedOnce cap} maybe necessary to ensure that the generated problem represents the task meaningully.  Through these experiments, we seek to answer the following questions:  

1) Given an expert demonstration and a natural language task description, can PDDL-ART generate syntactically valid PDDL domain and problem files? (\textit{\textbf{PDDL generation ability}})

2) Can VLMs generate PDDL domain, problem files that are semantically correct and enable generating a valid plan for a variety of tasks given natural language information and accompanying task setup images? (\textit{\textbf{High-level plan feasibility}})

3) Can VLMs evaluate predicates by reasoning about whether a predicate is perceptible in vision? If not, can the VLMs invoke appropriate tools and autonomously infer the input parameters to the tools, and use their output to evaluate the predicates? (\textit{\textbf{predicate evaluation}})

 4) Does iterative correction of the generated PDDL at various stages lead to PDDL descriptions that are semantically aligned with the expert demo? (\textit{\textbf{corrective performance}})

We use a reasoning VLM \textsc{GPT-O3} (model checkpoint - June 2026) and the most recent general-purpose VLM \textsc{GPT-5.5} (model checkpoint- June 2026) for our case studies.  Below we describe the experimental setup.

\subsection{Task scenarios}

\subsubsection{Engine: Check oil levels (easy)} The robot starts from an initial position away from the engine. The success criteria involves that the robot has inspected the oil levels and the dipstick has been inserted back into the engine. The task involves two objects: gripper and dipstick, and an efficient plan takes 5 actions to complete the task. 

\subsubsection{Engine: Open and close oil cap (medium)}  The robot starts from an initial position away from the engine. The success criteria is that the robot has unscrewed the oil cap, moved it away from the engine, and screwed it back onto the cap holder. The task involves two objects: gripper and cap, and an efficient plan takes 6 actions to complete the task.

\subsubsection{Engine: Change oil filter (medium)}  The robot starts from an initial position away from the engine, with the old oil filter attached to the engine and the new oil filter located to the side. The success criteria is that the old oil filter has been removed and the new oil filter has been installed on the engine in its place. The task involves three objects: gripper, old\_filter, and new\_filter, and an efficient plan takes 9 actions to complete the task.

 \subsubsection{Engine: Oil Change (hard)}  The robot starts from an initial position away from the engine, with the oil cap installed on the engine and the oil container on a side table The success criteria involves that oil has been poured into the engine and the oil cap has been reinstalled. The task involves three objects: gripper, cap, and oil\_container, and an efficient plan takes 11 actions to complete the task.

 \subsubsection{Household: Cooking Preparation I (medium)}  Starting from an initial position with an empty gripper, the robot is tasked to retrieve objects from inside fridge, cabinet and an open shelf. The task involves 5 objects and an efficient plan takes 17 steps to complete the task.

 \subsubsection{Household: Cooking Preparation II (hard)}  Same setup as above. This task involves 12 objects and the efficient plan takes 37 steps to complete the task.

\begin{table*}[t]
\centering
\caption{\centering \small Feasibility and success rate (\%) comparison of PDDL-ART with baselines 
across engine and household domains across 10 trials.}
\label{tab:success_rate}
\setlength{\tabcolsep}{4pt}
\renewcommand{\arraystretch}{1.4}
\resizebox{\textwidth}{!}{
\begin{tabular}{l | ccc | ccc | ccc | ccc |  ccc | ccc | ccc}
\toprule
\multirow{2}{*}{Method} 
  & \multicolumn{3}{c|}{Check Oil Levels} 
  & \multicolumn{3}{c|}{Cap Manipulation} 
  & \multicolumn{3}{c|}{Filter Change} 
  & \multicolumn{3}{c|}{Oil Change} 
  & \multicolumn{3}{c|}{Cooking Prep I} 
  & \multicolumn{3}{c|}{Cooking Prep II} 
  & \multicolumn{3}{c}{Average} \\
 & F $\uparrow$ & T $\downarrow$ & C $\uparrow$
 & F $\uparrow$ & T $\downarrow$ & C $\uparrow$ 
 & F $\uparrow$ & T $\downarrow$ & C $\uparrow$ 
 & F $\uparrow$ & T $\downarrow$ & C $\uparrow$ 
 & F $\uparrow$& T $\downarrow$ & C $\uparrow$ 
 & F $\uparrow$& T $\downarrow$ & C $\uparrow$
  & F $\uparrow$ & T $\downarrow$ & C $\uparrow$\\
\midrule \hline

Direct$_{\text{GPT-O3}}$   
  & \textbf{100} & 0 & \textbf{100}  & \textbf{100} & 0 &\textbf{ 100} & 10 & 10 & 0& 60& 0& 20& 70& 0& 50 & 60 & 0 & 50 & 66.67 & 1.67 & 53.33\\ 
Direct$_{\text{GPT-5.5}}$    
 &\textbf{ 100 }& 0 &\textbf{100} &\textbf{100 }& 0 &\textbf{100}& 20 & 10 & 10 & 70& 10& 40& 80 & 0 & 70  & 80 & 0 & 80 & 75 & 3.33 & 66.67\\ 
CoT$_{\text{GPT-O3}}$      
  & \textbf{100} & 0 & \textbf{100}& \textbf{100} & 0 &\textbf{100} &  80 & 0 & 10& 60& 0& 60&  80 & 0&80 & 80 & 0 & 70 & 83.33 & 0 &  70\\ 
CoT$_{\text{GPT-5.5}}$       
  & \textbf{100} & 0 & \textbf{100}& \textbf{100} & 0 & \textbf{100} &  {90}&  0 & 30&80 & 0& 60& \textbf{ 100} &0  & \textbf{100}& 80  & 0 & 80 & 91.67& 0 & 78.33\\  
NL2PLAN$_{\text{GPT-O3}}$ 
  & 80 & 0 & 20&  70 & 20 & 20 & 40 & 0 & 0 & 60 & 0 & 0&  20 & 0& 0 & 10 & 0 & 0 & 46.67 & 3.33 & 6.67\\ 
\hline 

 {PDDL-ART}$_{\text{GPT-O3}}$ \scriptsize{zero-shot} 
  &90  & 0  & 80& \textbf{100} & 40 & 60 & \textbf{100} & 0 & 0& 50 & 30 & 0& 40& 0 & 10 & 30 & 0 & 20 & 56.83 & 11.67 & 28.33 \\  

 {PDDL-ART}$_{\text{GPT-O3}}$ \scriptsize{w/ stage-1}
  &\textbf{100}  & 0  &\textbf{100}& \textbf{100} & 0 & 80 &\textbf{100 }& 0 & 0& \textbf{90 }& 10 & 0& 60& 0 & 20 & 80 & 0 & 50 & 88.33 & 1.67 & 41.67\\  

 {PDDL-ART}$_{\text{GPT-5.5}}$ \scriptsize{zero-shot} 
  &\textbf{100}  & 0  &\textbf{ 100} & \textbf{100} & 0 & \textbf{100} & 80 & 0 & 50& 50 & 30 & 0&  90& 0 & 20 & 70 & 0 & 60 & 81.62 & 5 &55\\  

 {PDDL-ART}$_{\text{GPT-5.5}}$ \scriptsize{w/ stage-1}
  &\textbf{100 } & 0  & \textbf{100}&\textbf{100} & 0 & \textbf{100} & 90 & 0 & 80& \textbf{90} & 10 & 30& \textbf{100}& 0 & 90 & \textbf{100} & 0 & 60 & \textbf{96.67} & 1.67 & 76.67\\

\textbf{PDDL-ART}$_{\textbf{GPT-O3}}$ 
  & \textbf{100 }& 0  & \textbf{100} & \textbf{100} & 0 & 90 & \textbf{100} & 0 & 60& \textbf{90}& 0& 70&  80 & 0& 80 & 80 & 0 & 80 & 91.62 & 0 & 80\\

  \textbf{PDDL-ART}$_{\textbf{GPT-5.5}}$ 
  &\textbf{100}  & 0  &\textbf{100}& \textbf{100} & 0 & \textbf{100} & 90 & 0 & \textbf{80}& \textbf{90} & 0 & \textbf{90}&  \textbf{100}& 0 & \textbf{100} & 90 & 0 & \textbf{90} & {95} & 0 & \textbf{93.33}\\  
\bottomrule
\end{tabular} }
\end{table*}

\subsection{Baselines}

 {While related approaches have been proposed recently, such as \cite{hao2025simulation}, they require fine-tuning a VLM on specific domains to enable spatial reasoning. Since we do not assume access to such fine-tuning data for our method, we do not compare to these methods. Instead we compare to the following methods. 

 }

\subsubsection{VLM-PLAN} The VLM is provided with the task demonstration images, 
task description, and action library, and directly generates an action sequence. 
\subsubsection{VLM-PLAN (CoT)}Same setup as VLM-PLAN, but the VLM is additionally prompted 
to perform chain-of-thought reasoning prior to outputting the action sequence.
 
\subsubsection{NL2PLAN~\cite{gestrin-et-al-icaps2024wshaxp}} An LLM-based approach that iteratively generates PDDL problem and domain files. The approach performs $N$ iterations at every corrective stage and uses 6 stages, each validating one aspect of PDDL - such as extracting type hierarchies, formulating action schema, etc. We limit the iterations at every step to 3. The approach outputs domain and problem files. We use FastDownward~\cite{helmert2006fast} for synthesizing plans on the PDDL descriptions generated by NL2PLAN and PDDL-ART.

\subsection{Evaluation metrics}

1. Planning feasibility (F) - Syntactically valid descriptions may still be incorrect - the generated domain should have preconditions and effects correctly defined that can be sequenced together to go from initial state to goal - if there are missing links, planning maybe infeaisble. We treat trivial plans as a special case - these are empty plans but hint at incorrect init goal conditions for the problem. While planning feasibility still considers this as a valid instance, we highlight the values in parenthesis next to the planning feasibility result if number of trivial plans generated in non-zero, 

2. Number of trivial plans - whether or not the generated plans are 0-length. This metric reflects incorrect modeling of the problem where the goal is trivially satisfied.

3. Plan alignment (C) - This post-hoc metric reflects if the executed plans are semantically aligned with the task demonstration. Note that a plan may be inefficient, but still correct as it may involve  redundant actions without affecting the correctness and executability of the plan. This metric counts inefficient plans as success. It is important to note that this metric is only used for evaluation and is not available for any correction stage in PDDL-ART. 

\subsection{Results and Analysis}

Table~\ref{tab:success_rate} shows the performance of all baseline approaches and our method across six tasks and ten trials. The Direct planning approach performs well on easy tasks however its accuracy of feasible and aligned plans decreases with complexity. This is because missing even one action in the plan may render it infeasible. The CoT method shows improvement over Direct planning due to explicit reasoning requirements. The performance of NL2PLAN across all tasks shows that while similar iterative correction approaches can be used, the visual information used in our method likely plays a key role in guiding the models towards valid and aligned PDDL descriptions. Our method performs as well as the direct and CoT plan baselines on easy tasks, and outperforms them on more complex, long-horizon tasks, achieving an overall feasibility rate of 95\% and a 93.3\% plan alignment rate.

\subsection{Effectiveness of various components of PDDL-ART} 

We perform an ablation analysis of effectiveness of each stage of our method. We consider zero-shot PDDL generation, PDDL-ART with only stage-1 corrections, and the overall proposed framework. Table~\ref{tab:success_rate} shows that the while PDDL-ART zero-shot achieves almost perfect compilation success and planning feasibility on easy tasks, it struggles to generate valid PDDL as the tasks become more complex. PDDL-ART w/stage-1 shows consistent improvement in terms of compilation success over zero-shot generation across all tasks, supporting the hypothesis that syntactic and semantic validation-guided correction may improve the generated PDDL. However, the plan alignment metric shows that, while this stage may assist in creating valid and feasible PDDL, the resulting plans may not be executable or may not be aligned with the demonstration. Finally, the overall PDDL-ART framework achieves high success rate both in terms of number of plans found and the task alignment metric across all tasks after benefitting from both stage-1 and stage-2 corrections. Note that the corrective processes may introduce modifications that can make the problem infeasible, as evident from the results for Cooking Prep II task.

\section{Discussion and Conclusion}



We present \textsc{PDDL-ART}, a framework for autonomously generating and validating PDDL domain and problem descriptions from a single task demonstration and natural language description, without requiring domain templates, action signatures, or fine-tuning. The multi-stage correction pipeline systematically addresses syntactic, semantic, and execution-level errors, while the tool-augmented predicate evaluation grounds symbolic verification in geometric and temporal reasoning. Together, these components reduce reliance on domain expertise while producing PDDL descriptions that are aligned with the demonstrated task. 

The proposed framework generates domain descriptions scoped to the provided demonstration, i.e., the resulting domain contains only the predicates and actions inferred from the demo, rather than a general domain supporting diverse task instances from a domain. Extending the framework to generate reusable domains across a broader set of task instances is a direction for future work. The framework also currently assumes fully observable, deterministic planning and does not support conditional, temporal, or numeric planning constructs, which are natural next steps. 

Finally, while reducing dependence on domain experts is a key strength of our framework, it also introduces a limitation: VLM-guided predicate evaluation may be prone to hallucinations or visual noise, potentially leading to unnecessary corrections or, in worse cases, introducing errors into previously valid parts of the PDDL description.

\bibliographystyle{IEEEtran}
\bibliography{references}

\end{document}